# Deep Neural Networks based Modrec: Some Results with Inter-Symbol Interference and Adversarial Examples

S. Asim Ahmed
Advanced Technology Center
*Rockwell Collins*
Cedar Rapids, USA
asim.ahmed@rockwellcollins.com

Subhashish Chakravarty
Advanced Technology Center
*Rockwell Collins*
Cedar Rapids, USA
subhashish.chakravarty@rockwellcollins.com

Michael Newhouse
Advanced Technology Center
*Rockwell Collins*
Cedar Rapids, USA
michael.newhouse@rockwellcollins.com

*Abstract*—Recent successes and advances in Deep Neural Networks (DNN) in machine vision and Natural Language Processing (NLP) have motivated their use in traditional signal processing and communications systems. In this paper, we present results of such applications to the problem of automatic modulation recognition. Variations in wireless communication channels are represented by statistical channel models and their parameterization will increase with the advent of 5G. In this paper, we report effect of simple two path channel model on our naive deep neural network based implementation. We also report impact of adversarial perturbation to the input signal.

## I. INTRODUCTION

The problem of reliably and efficiently classifying modulation type and its arity is of important practical significance with increasing diffusion of wireless communications. An efficient and robust solution to the problem may be useful for spectrum monitoring, cooperative spectrum sharing for cognitive radios and various other military applications including Low Probability of Intercept (LPI) waveforms.

Various approaches to modulation recognition emphasize different aspects of the ML pipeline: 1) feature extraction (hand-crafted features, automated features or hybrid), 2) processing algorithms (SVM, GLM, decision tree, neural network) and 3) decision framework (traditional vs Bayesian approach). Although there is a continuing healthy debate amongst these approaches, we will focus on a simple all neural network solution in this paper. We will also build in some simplifying assumptions into our data with fixed length input data format (128 complex samples).

In this paper, we share experimental results showing impact of a simple two-path channel model and adversarial perturbations on the accuracy of a convolutional neural network based modulation recognition architecture [1]. While adversarial examples have attracted a lot of attention in the machine learning community, one of our goals is to raise awareness about robustness/fragility of current ML algorithms to Inter-Symbol Interference (ISI) and adversarial examples in signal processing and communications community.

## II. EXPERIMENTAL SETUP – DATA USED FOR THE PAPER

Data was generated with GNU radio script "generate_RML2016.10a.py" located at /radioML/dataset on Github provided by [1]. The script generates 11 types of modulations at Signal to Noise Ratios (SNRs) ranging from -20dB to +18dB in steps of 2dB. For our training, we excluded signals with SNR less than -4dB. Each modulation and SNR combination contains 1000 waveforms of 128 complex samples. More details about dataset may be found in [1].

We derived 3 separate datasets from the data provided by [1].

- dataset1: Additive White Gaussian Noise (AWGN) data as above.

- dataset2: The GNU radio script was modified to generate 144 samples that are replicated, delayed and scaled by a complex number (0.2781,0.856) to create a specular path and then added back to the original 144 samples. These samples are truncated to 128 samples representing two-path ISI multipath signal. The specular path is further scaled by 0.25, 0.5 and 1.0 to create multi-path profile with varying intensity. This data is used for testing multi-path performance of various trained networks.

- dataset3: Similar to dataset2, except that the path delay is randomized uniformly over the second symbol. This data was used for training ISI neural networks.

## III. DEEP CONVOLUTIONAL NEURAL NETWORK ARCHITECTURE

We followed the model of a neural network structure given in [1] while changing the number of hidden units, learning rate and L2 regularization coefficients. We used raw samples as input with and without scaling and did not add any hand-crafted features like higher order moments (cumulants) or spectral transformations for this paper. We also used dropout of 60%. The number of layers and their types was kept same for all training. The network is composed of input, convolutional, convolutional, fully connected, fully connected and softmax layers. The number of hidden units in all layers are part of hyper parameters to be tuned by validation data. Fig. 1 shows distribution of accuracy over 80 different hyper-parameters of the neural networks. The hyper-parameters also include learning rate and regularization coefficients. The figure shows reasonably good correlation between training and validation accuracy. We followed Xavier initializations for Convolutional layers and Glorot/He [3][4] (normally distributed) initializations for fully connected layers. We used Adam stochastic optimization method [5] for all trials.

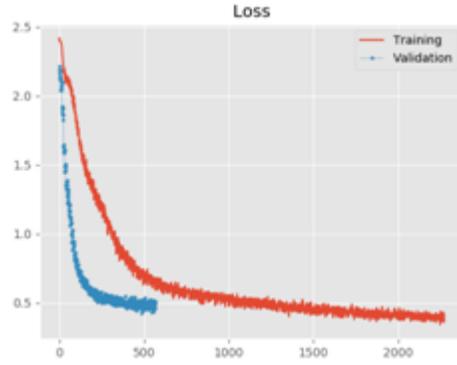

Fig. 3 Loss metric for each batch of training and validation data over the training process

## A. Additive White Gaussian Noise

Fig. 4 shows confusion matrix of DNN based modulation recognition with AWGN. There seems to be significant ambiguity between (QAM16, QAM64) and (AM-DSB, WBFM) modulations.

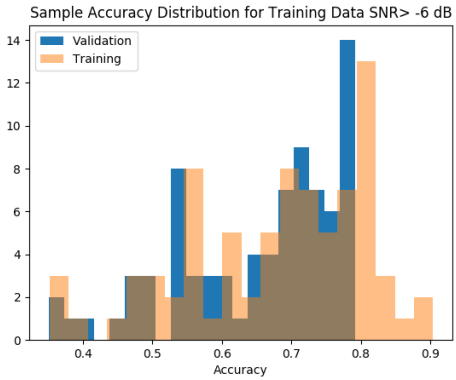

Fig. 1 Sample Accuracy of Training and Validation over 80 combinations of hyper-parameters

Fig. 2 shows the network architecture and its parameterization derived from dataset1(AWGN) training. The network was chosen based on its performance and size (to avoid overfitting). The overall accuracy is 78% on test data set.

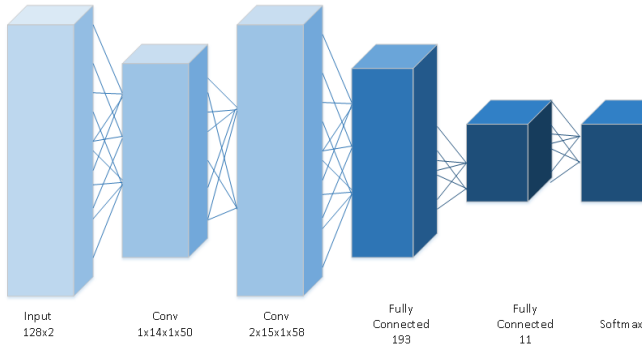

Fig. 2 Network Architecture

Fig. 3 shows the plot of training and test loss metric over each batch in the training process. The loss metric is calculated as cross entropy of known classification and softmax output of the network.

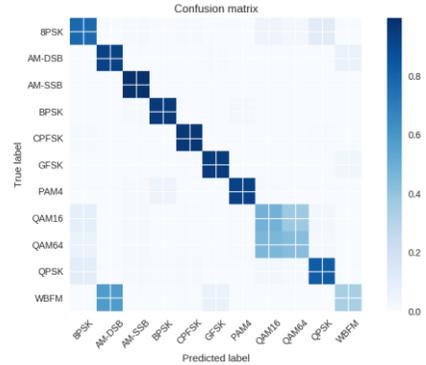

Fig. 4 Performance of selected network

## B. Inter-Symbol Interference

The network trained on dataset1 (AWGN data) then was evaluated with dataset2 (ISI data). The overall performance (scale=1.0) decreased to 19% from 76% (AWGN). Fig. 5, Fig. 6, and Fig. 7 shows the classification accuracy with decreasing levels of ISI. The figures show that the performance improves, as expected, as the scale/intensity of second path is lowered.

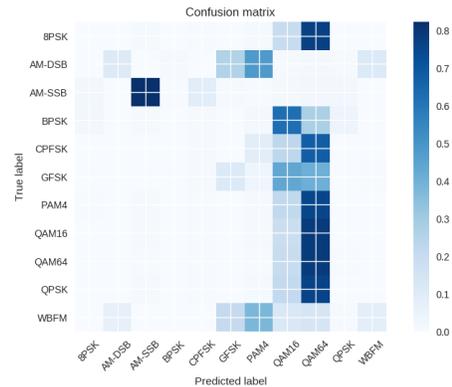

Fig. 5 ISI Accuracy with scale = 1.0, Test Accuracy = 0.193

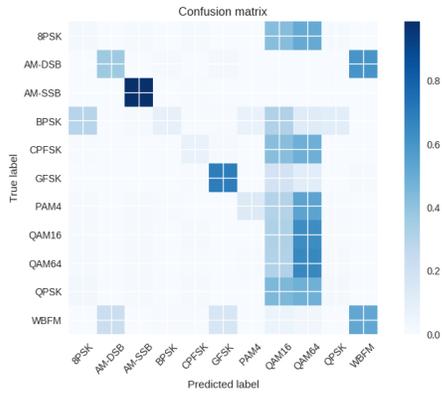

Fig. 6 ISI Accuracy with scale = 0.5, Test Accuracy = 0.351

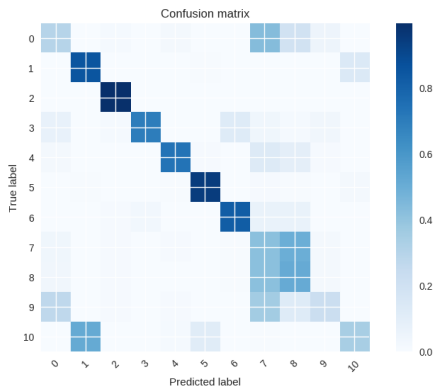

Fig. 7 ISI Accuracy with scale = .25, Test Accuracy = 0.625

Various networks were trained with dataset3 (two-path data with random delayspread). The test accuracy improved to 25% from 19%. It also resulted in both lowered training accuracy and overfitting. Test samples generated from dataset3 performed significantly better (64%) than samples from dataset2 indicating significant overfitting (25%). Fig. 8 and Fig. 9 show the confusion matrix for test samples from dataset3 and dataset2.

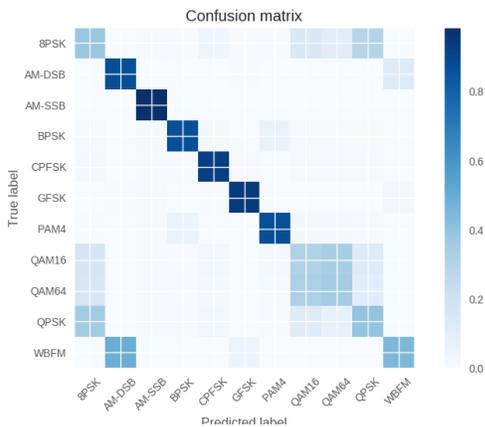

Fig. 8 ISI Training (dataset3) – Test (dataset3) Accuracy 64%

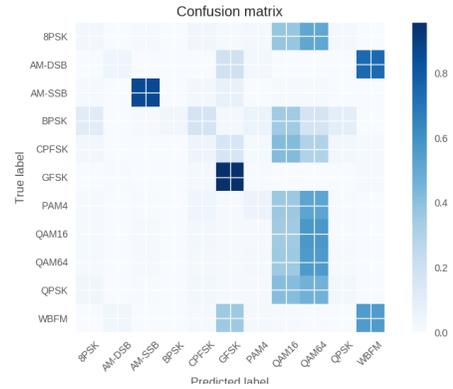

Fig. 9 ISI Training (dataset3)-Test (dataset2) Accuracy 25%

TABLE I summarizes the modulation results for AWGN and two path channel models.

**TABLE I. ISI Result Summary**

| Training data | Test Data 1 | Accuracy |
|---|---|---|
| Dataset1 / AWGN | Dataset1 / AWGN | 78% |
| Dataset1 / AWGN | Dataset2 / ISI | 19% |
| Dataset3 / ISI | Dataset3 / ISI | 64% |
| Dataset3 / ISI | Dataset2 / ISI | 25% |

*C. Adversarial Examples*

Deep neural network based machine learning algorithms have had tremendous success in machine vision. These algorithms have out-performed humans on natural images. Surprisingly, recent adversarial examples have shown that these high performances are not robust to adversarial perturbations. In their seminal work [2], the authors have identified linearity in high dimensional space as the cause of adversarial vulnerability. They have also identified tension between ease in training (linearity) and performance with adversarial examples. In this paper, we use the fast gradient sign method [2] to generate adversarial perturbations according to the equation below

$$\eta = \epsilon\, sign(\nabla_x J(\theta, x, y))$$

The epsilon controls max norm of the perturbation. It is instructive to note that for a given value of epsilon, L2 norm of corresponding perturbations grow linearly with dimension of the problem.

For adversarial examples, we used dataset1 exclusively. Fig. 10 and Fig. 11 show two examples of adversarial perturbation (with epsilon=0.05) that change the classification. Fig. 10 shows that adversarial perturbation flips an incorrectly classified signal to correct classification and Fig. 11 shows more typical case of adversarial perturbation leading to misclassification. More importantly, in both cases the confidence level of the resulting adversarial prediction is very high (~0.8 and 0.9) for multinomial classification which is disconcerting given that signals themselves look almost identical.

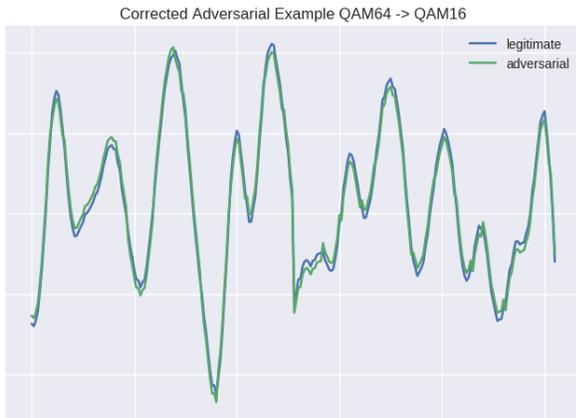

Fig. 10 Legitimate and Adversarial (high SNR) example waveform

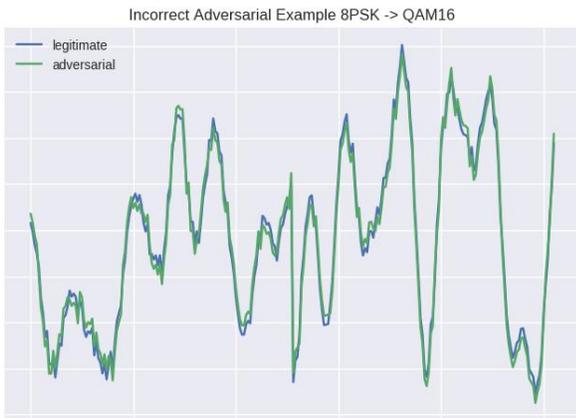

Fig. 11 Legitimate and Adversarial (high SNR) example waveform

Fig. 12 shows the plot of accuracy against various levels of epsilon and Fig. 13 shows comparison legitimate vs adversarial accuracy over SNR.

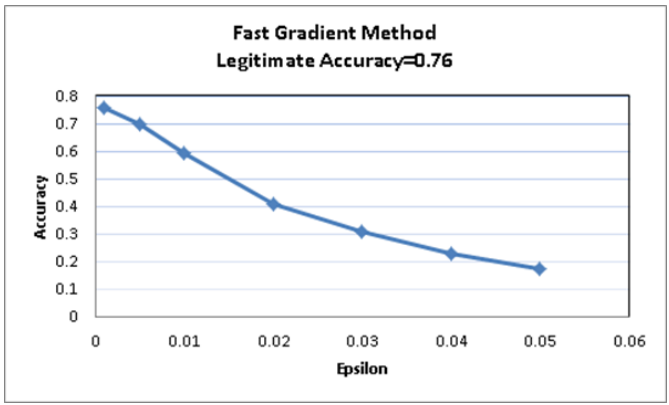

Fig. 12 Performance degradation with increasing value of Epsilon

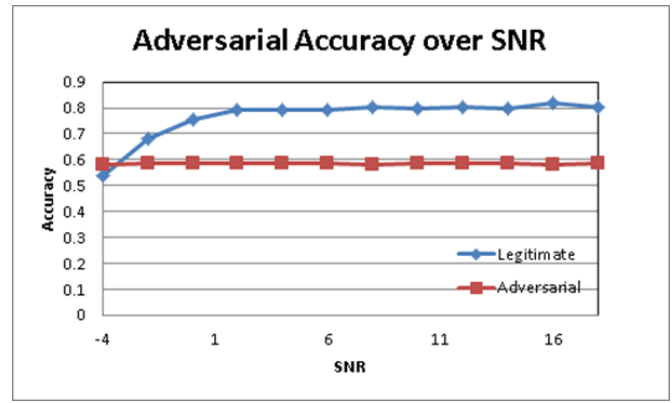

Fig. 13 Non-Adversarial and Adversarial Accuracy Comparison

### D. Input Scaling

We observe that performance results are sensitive to input scaling. For example, test accuracy improved to 94% with use of scaling oracle. The results suggest that there may be ways to improve performance with a preprocessing neural network. However, we also note that improper scaling degraded performance. Multipath signals can exhibit significantly wider dynamic range compared to the transmitted waveform and it is unclear how to optimally scale input signals. Training with higher SNR signals (>8dB) only did not result in significant improvement of accuracy performance.

## IV. CONCLUSION

Deep neural networks typically use linear optimization techniques to learn parameters in extremely high dimensions and most activations in the network are carefully controlled during the training process to remain in the linear region due to vanishing gradient problem. We see that this leads to sensitivity to input scaling and the adversarial perturbation problem. Also, we saw that our simple network performs poorly in multi-path environment. Both multi-path and adversarial examples point to more complex networks and optimization techniques to mitigate their effects. Future work may include exploration of encoder-decoder style network like variational autoencoders to model scaling and ISI as latent variables.